\pdfoutput=1
\documentclass[11pt]{article}

\usepackage{multirow}
\usepackage[table,xcdraw]{xcolor}

\usepackage[]{emnlp2021}

\usepackage{times}
\usepackage{latexsym}

\usepackage[T1]{fontenc}

\usepackage[utf8]{inputenc}

\usepackage{microtype}

\usepackage{here}
\usepackage{latexsym}
\usepackage{booktabs}
\usepackage{stfloats}
\usepackage{subcaption}
\usepackage[T5,T1]{fontenc}

\usepackage{amsmath}
\usepackage{amsfonts}
\usepackage{cleveref}
\usepackage{float}

\usepackage{todonotes}
\usepackage{cancel}
\usepackage{xspace}
\usepackage{amsthm}
\usepackage{amssymb}
\usepackage{comment}
\usepackage{adjustbox}
\usepackage{tipa}
\usepackage{enumitem}
\usepackage{textcomp}

\title{How Familiar Does That Sound? Cross-Lingual Representational Similarity Analysis of Acoustic Word Embeddings}

\author{Badr M. Abdullah  \hspace{0.75cm} Iuliia Zaitova  \hspace{0.75cm} Tania Avgustinova \\
  \textbf{Bernd Möbius \hspace{0.75cm} Dietrich Klakow } \\ 
         Department of Language Science and Technology (LST)  \\ Saarland Informatics Campus, Saarland University, Germany \\ 
        Corresponding author: \normalsize{\textsf{babdullah@lsv.uni-saarland.de}} 
}

\begin{document}
\maketitle
\begin{abstract}

How do neural networks ``perceive'' speech sounds from unknown languages? Does the typological similarity between the model's training language (L1) and an unknown language (L2) have an impact on the model representations of L2 speech signals? To answer these questions, we present a novel experimental design based on representational similarity analysis (RSA) to analyze acoustic word embeddings (AWEs)---vector representations of variable-duration spoken-word segments. First, we train monolingual AWE models on seven Indo-European languages with various degrees of typological similarity. We then employ RSA to quantify the cross-lingual similarity by simulating native and non-native spoken-word processing using AWEs. Our experiments show that typological similarity indeed affects the representational similarity of the models in our study. We further discuss the implications of our work on modeling speech processing and language similarity with neural networks.

\end{abstract}

\section{Introduction}
Mastering a foreign language is a process that requires (human) language learners to invest time and effort. If the foreign language (L2) is very distant from our native language (L1), not much of our prior knowledge of language processing would be relevant in the learning process. On the other hand, learning a language that is similar to our native language is much easier since our prior knowledge becomes more useful in establishing the correspondences between L1 and L2 \citep{ringbom2006cross}. In some cases where L1 and L2 are closely related and typologically similar, it is possible for an L1 speaker to comprehend L2 linguistic expressions to a great degree without prior exposure to L2. The term \textit{receptive multilingualism} \citep{zeevaert2007receptive} has been coined in the sociolinguistics literature to describe this ability of a listener to comprehend utterances of an unknown but related language without being able to produce it.

Human speech perception has been an active area of research in the past five decades which has produced a wealth of documented behavioral studies and experimental findings. Recently, there has been a growing scientific interest in the cognitive modeling community to leverage the recent advances in speech representation learning to formalize and test theories of speech perception using computational simulations on the one hand, and to investigate whether neural networks exhibit similar behavior to humans on the other hand \cite{Rsnen2016AnalyzingDL, alishahi-etal-2017-encoding, DUPOUX201843, scharenborg2019representation,  gelderloos-etal-2020-learning, MatusevychSKFG20, magnuson2020earshot}. 

In the domain of modeling non-native speech perception,  \citet{schatz2018neural} have shown that a neural speech recognition system (ASR) predicts Japanese speakers' difficulty with the English phonemic contrast /\textipa{l}/-/\textipa{\*r}/ as well as English speakers' difficulty with the Japanese vowel length distinction. \citet{matusevych2021phonetic} have shown that a model of non-native spoken word processing based on neural networks predicts lexical processing difficulty of English-speaking learners of Russian. The latter model is based on word-level representations that are induced from naturalistic speech data known in the speech technology community as acoustic word embeddings (AWEs). AWE models map variable-duration spoken-word segments onto fixed-size representations in a vector space such that instances of the same word type are (ideally) projected onto the same point in space \cite{levin+etal_asru13}. In contrast to word embeddings in natural language processing (NLP), an AWE encodes information about the acoustic-phonetic and phonological structure of the word, not its semantic content.

\begin{figure*}
    \centering
    \includegraphics[width=0.90\textwidth]{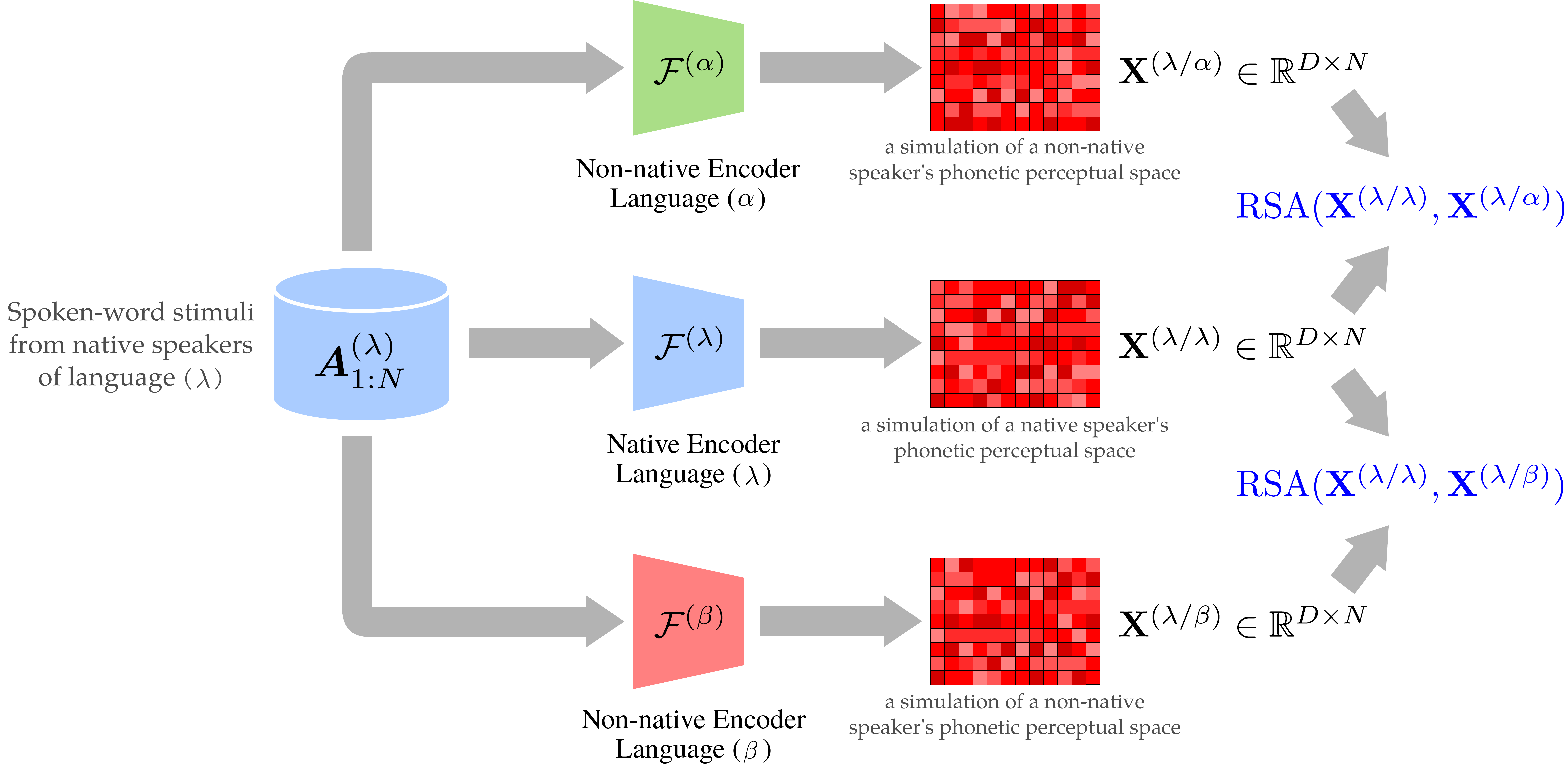}
    \caption{An illustrated example of our experimental design. A set of $N$ spoken-word stimuli from language $\lambda$ are embedded using the encoder $\mathcal{F}^{(\lambda)}$ which was trained on language $\lambda$ to obtain a (native) view of the data: $\mathbf{X}^{(\lambda/\lambda)} \in \mathbb{R}^{D \times N}$. Simultaneously, the same stimuli are embedded using encoders trained on other languages, namely $\mathcal{F}^{(\alpha)}$ and $\mathcal{F}^{(\beta)}$, to obtain two different (non-native) views of the data: $\mathbf{X}^{(\lambda/\alpha)}$ and $\mathbf{X}^{(\lambda/\beta)}$. We quantify the cross-lingual similarity between two languages by measuring the association between their embedding spaces using the representational similarity analysis (RSA) framework. } 
    \label{fig:RSA_view}
\end{figure*}

Although the model of  \citet{matusevych2021phonetic} has shown similar effects to what has been observed in behavioral studies \cite{cook2016fuzzy}, it remains unclear to what extent AWE models can predict a facilitatory effect of language similarity on cross-language spoken-word processing. In this paper, we present a novel experimental design to probe the \textit{receptive multilingual} knowledge of monolingual AWE models (i.e., trained without L2 exposure) using the representational similarity analysis (RSA) framework. In a controlled experimental setup, we use AWE models to simulate spoken-word processing of native speakers of seven languages with various degrees of typological similarity. We then employ RSA to characterize how language similarity affects the emergent representations of the models when tested on the same spoken-word stimuli (see Figure~\ref{fig:RSA_view} for an illustrated overview of our approach). Our experiments demonstrate that neural AWE models of different languages exhibit a higher degree of representational similarity if their training languages are typologically similar.

\section{Proposed Methodology}

A neural AWE model can be formally described as an encoder function $\mathcal{F}: \mathcal{A} \xrightarrow[]{} \mathbb{R}^D$, where $\mathcal{A}$ is the (continuous) space of acoustic sequences and $D$ is the dimensionality of the embedding. Given an acoustic word signal represented as a temporal sequence of $T$ acoustic events $\boldsymbol{A} = (\boldsymbol{a}_1, \boldsymbol{a}_2, ..., \boldsymbol{a}_T)$, where $\boldsymbol{a}_t \in \mathbb{R}^{k}$ is a spectral vector of $k$ coefficients, an embedding is computed as
\begin{equation}
\mathbf{x} = \mathcal{F}(\boldsymbol{A}; \boldsymbol{\theta}) \in \mathbb{R}^D
\end{equation}
Here, $\boldsymbol{\theta}$ are the parameters of the encoder, which are learned by training the AWE model in a monolingual supervised setting. That is, the training spoken-word segments (i.e., speech intervals corresponding to spoken words) are sampled from utterances of native speakers of a single language where the word identity of each segment is known. The model is trained with an objective that maps different spoken segments of the same word type onto similar embeddings. To encourage the model to abstract away from speaker variability, the training samples are obtained from multiple speakers, while the resulting AWEs are evaluated on a held-out set of speakers. 

Our research objective in this paper is to study the discrepancy of the representational geometry of two AWE encoders that are trained on different languages when tested on the same set of (monolingual) spoken stimuli. To this end, we train AWE encoders on different languages where the training data and conditions are comparable across languages with respect to size, domain, and speaker variability. We therefore have access to several encoders $\{\mathcal{F}^{(\alpha)}, \mathcal{F}^{(\beta)}, \dots, , \mathcal{F}^{(\omega)}\}$, where the superscripts $\{\alpha, \beta, \dots, \omega\}$ denote the language of the training samples.

Now consider a set of $N$ held-out spoken-word stimuli produced by native speakers of language $\lambda$: $\boldsymbol{A}_{1:N}^{(\lambda)} = \{\boldsymbol{A}_1^{(\lambda)}, \dots, \boldsymbol{A}_N^{(\lambda)}\}$. First, each acoustic word stimulus in this set is mapped onto an embedding using the encoder $\mathcal{F}^{(\lambda)}$, which yields a matrix $\mathbf{X}^{(\lambda/\lambda)} \in \mathbb{R}^{D \times N}$. Since the encoder $\mathcal{F}^{(\lambda)}$ was trained on language $\lambda$, we refer to it as the \textit{native encoder} and consider the matrix $\mathbf{X}^{(\lambda/\lambda)}$ as a simulation of a native speaker's phonetic perceptual space. To simulate the phonetic perceptual space of a non-native speaker, say a speaker of language $\alpha$, the stimuli $\boldsymbol{A}_{1:N}^{(\lambda)}$ are embedded using  the encoder $\mathcal{F}^{(\alpha)}$, which yields a matrix $\mathbf{X}^{(\lambda/\alpha)} \in \mathbb{R}^{D \times N}$. Here, we read the superscript notation $(\lambda/\alpha)$ as word stimuli of language $\lambda$ encoded by a model trained on language $\alpha$.  Thus, the two matrices $\mathbf{X}^{(\lambda/\lambda)}$ and $\mathbf{X}^{(\lambda/\alpha)}$ represent two different views of the same stimuli. Our main hypothesis is that the cross-lingual representational similarity between emergent embedding spaces should reflect the  acoustic-phonetic and phonological similarities between the languages $\lambda$ and $\alpha$. That is, the more distant languages $\lambda$ and $\alpha$ are, the more dissimilar their corresponding representation spaces are. To quantify this cross-lingual representational similarity, we use Centered Kernel Alignment (CKA) \citep{kornblith2019similarity}. CKA is a representation-level similarity measure that emphasizes the distributivity of information and therefore it obviates the need to establish the correspondence mapping between single neurons in the embeddings of two different models. Moreover, CKA has been shown to be invariant to orthogonal transformation and isotropic scaling which makes it suitable for our analysis when comparing different languages and learning objectives. Using CKA, we quantify the similarity between languages $\lambda$ and $\alpha$ as   
\begin{equation}
\textsf{sim}(\lambda, \alpha) := \textsf{CKA}(\mathbf{X}^{(\lambda/\lambda)}, \mathbf{X}^{(\lambda/\alpha)})
\end{equation}
Here, $\textsf{sim}(\lambda, \alpha) \in [0, 1]$ is a scalar that measures the correlation between the responses of the two encoders, i.e., native  $\mathcal{F}^{(\lambda)}$ and non-native  $\mathcal{F}^{(\alpha)}$, when tested with spoken-word stimuli $\boldsymbol{A}_{1:N}^{(\lambda)}$. $\textsf{sim}(\lambda, \alpha) = 1$ is interpreted as perfect association between the representational geometry of models trained on languages $\lambda$ and $\alpha$ while $\textsf{sim}(\lambda, \alpha) = 0$ indicates that no association can be established.  Likewise, we obtain another non-native view $\mathbf{X}^{(\lambda/\beta)}$ of the stimuli $\boldsymbol{A}_{1:N}^{(\lambda)}$ using another non-native encoder $\mathcal{F}^{(\beta)}$. We then quantify the cross-lingual representational similarity between languages $\lambda$ and $\beta$ as   
\begin{equation}
\textsf{sim}(\lambda, \beta) := \textsf{CKA}(\mathbf{X}^{(\lambda/\lambda)}, \mathbf{X}^{(\lambda/\beta)})
\end{equation}
If $\textsf{sim}(\lambda, \alpha) > \textsf{sim}(\lambda, \beta)$, then we interpret this as an indication that the native phonetic perceptual space of language $\lambda$ is more similar to the non-native phonetic perceptual space of language $\alpha$, compared to that of language $\beta$. Note that while $\textsf{CKA}(.,.)$ is a symmetric metric (i.e., $\textsf{CKA}(\mathbf{X},\mathbf{Y}) = \textsf{CKA}(\mathbf{Y},\mathbf{X}) $), our established similarity metric $\textsf{sim}(.,.)$ is not symmetric (i.e., $\textsf{sim}(\lambda, \alpha) :\neq  \textsf{sim}(\alpha, \lambda)$). To  estimate $\textsf{sim}(\alpha, \lambda)$, we use word stimuli of language $\alpha$ and collect the matrices $\mathbf{X}^{(\alpha/\alpha)}$ and $\mathbf{X}^{(\alpha/\lambda)}$. Then we compute 
\begin{equation}
\textsf{sim}(\alpha, \lambda) := \textsf{CKA}(\mathbf{X}^{(\alpha/\alpha)}, \mathbf{X}^{(\alpha/\lambda)})
\end{equation}

When we apply the proposed experimental pipeline across $M$ different languages, the effect of language similarity can be characterized by constructing a cross-lingual representational similarity matrix (xRSM) which is an asymmetric $M \times M$ matrix where each cell represents the correlation (or agreement) between two embedding spaces.   

\section{Acoustic Word Embedding Models}

\begin{figure*}[t]
    \centering
    \includegraphics[width=0.99\textwidth]{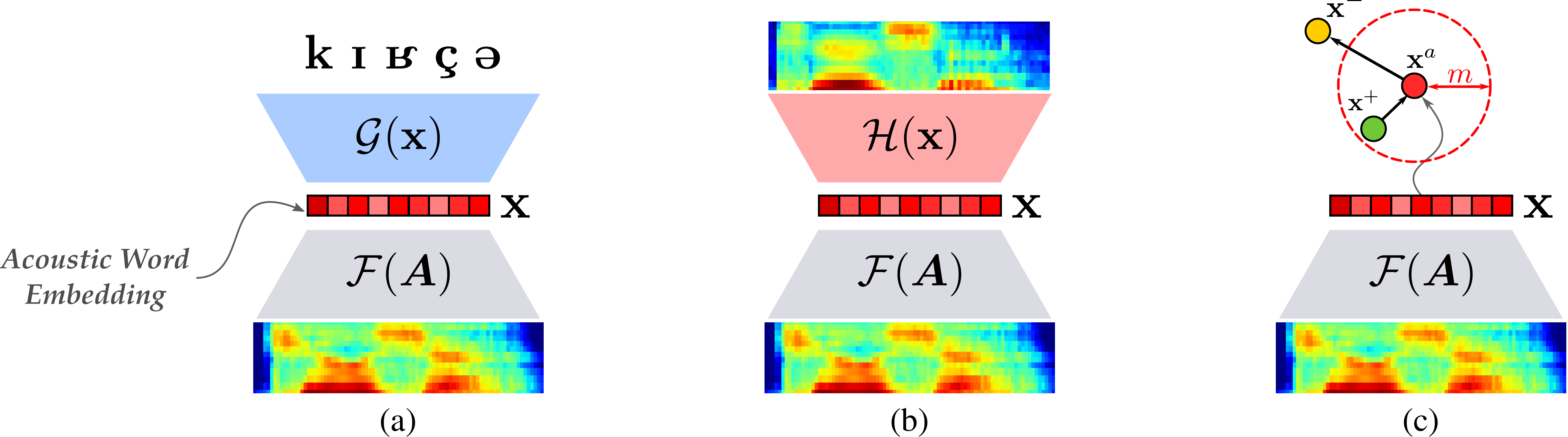}
    \caption{A visual illustration of the different learning objectives for training AWE encoders: (a) phonologically guided encoder (PGE): a sequence-to-sequence network with a phonological decoder, (b) correspondence auto-encoder (CAE): a sequence-to-sequence network with an acoustic decoder, and (c) contrastive siamese encoder (CSE): a contrastive network trained via triplet margin loss.}
    \label{fig:models}
\end{figure*}

We investigate three different approaches of training AWE models that have been previously introduced in the literature. In this section, we formally describe each one of them.
\subsection{Phonologically Guided Encoder}
The phonologically guided encoder (PGE) is a sequence-to-sequence model in which the network is trained as a word-level acoustic model \cite{abdullah21_interspeech}. Given an acoustic sequence ${\boldsymbol{A}}$ and its corresponding phonological sequence $\boldsymbol{\varphi} = (\varphi_1, \dots, \varphi_{\tau})$,\footnote{We use the phonetic transcription in IPA symbols as the phonological sequence $\boldsymbol{\varphi}$.} the acoustic encoder $\mathcal{F}$ is trained to take ${\boldsymbol{A}}$ as input and produce an AWE $\mathbf{x}$, which is then fed into a phonological decoder $\mathcal{G}$ whose goal is to generate the sequence $\boldsymbol{\varphi}$ (Fig.~\ref{fig:models}--a).  The objective is to minimize a categorical cross-entropy loss at each timestep in the decoder, which is equivalent to 
\begin{equation}
    \mathcal{L} = -  \sum_{i=1}^{\tau}{ \text{log } \mathbf{P}_{\mathcal{G}}(\varphi_{i} | \boldsymbol{\varphi}_{<i}, \mathbf{x})}
\end{equation}
where $\mathbf{P}_{\mathcal{G}}$ is the probability of the phone $\varphi_{i}$ at timestep $i$, conditioned on the previous phone sequence $\boldsymbol{\varphi}_{<i}$ and the embedding $\mathbf{x}$. The intuition of this objective is that spoken-word segments of the same word type would have identical phonological sequences, thus they are expected to end up nearby in the embedding space.

\subsection{Correspondence Autoencoder}

In the correspondence autoencoder (CAE) \cite{kamper2019truly}, each training acoustic segment ${\boldsymbol{A}}$ is paired with another segment that corresponds to the same word type ${\boldsymbol{A}^{+} = (\boldsymbol{a}_{1}^{+}, \boldsymbol{a}_{2}^{+}, ..., \boldsymbol{a}^{+}_{T^{+}}})$. The acoustic encoder $\mathcal{F}$ then takes  ${\boldsymbol{A}}$  as input and computes an AWE $\mathbf{x}$, which is then fed into an acoustic decoder $\mathcal{H}$ that attempts to reconstruct the corresponding acoustic sequence $\boldsymbol{A}^{+}$ (Fig. 2--b). The objective is to minimize a vector-wise mean square error (MSE) loss at each timestep in the decoder, which is equivalent to 
\begin{equation}
    \mathcal{L} = -  \sum_{i=1}^{T^{+}}{ ||\boldsymbol{a}_{i}^{+} - \mathcal{H}_i(\mathbf{x})||^2}
\end{equation}
where $\boldsymbol{a}_{i}^{+}$ is the ground-truth spectral vector at timestep $i$ and $\mathcal{H}_i(\mathbf{x})$ is the reconstructed  spectral vector at timestep $i$ as a function of the embedding $\mathbf{x}$. It has been hypothesized that learning the correspondence between two acoustic realizations of the same word type encourages the model to build up speaker-invariant word representations while preserving phonetic information \cite{matusevych2020analyzing}.

\subsection{Contrastive Siamese Encoder }
The contrastive siamese encoder (CSE) is based on a siamese network \cite{bromley1994signature} and has been explored in the AWEs literature with different underlying architectures \cite{settle+livescu_slt16, kamper+etal_icassp16}.  This learning objective differs from the first two as it explicitly minimizes/maximizes relative distances between AWEs of the same/different word types (Fig. 2--c). First, each acoustic word instance is paired with another instance of the same word type, which yields the tuple $(\boldsymbol{A}, \boldsymbol{A}^+)$. Given the AWEs of the instances in this tuple $(\mathbf{x}^a, \mathbf{x}^+)$, the objective is then to minimize a triplet margin loss
\begin{equation}
    \mathcal{L} = \text{max} \Big[0, m + d(\mathbf{x}^a, \mathbf{x}^+) - d(\mathbf{x}^a, \mathbf{x}^-) \Big]
\end{equation}
Here, $d: \mathbb{R}^D \times \mathbb{R}^D \rightarrow [0, 1]$ is the cosine distance and $\mathbf{x}^-$ is an AWE that corresponds to a different word type sampled from the mini-batch such that the term $d(\mathbf{x}^a, \mathbf{x}^-)$ is minimized. This objective aims to map acoustic segments of the same word type closer in the embedding space while pushing away segments of other word types by a distance defined by the margin hyperparameter $m$.

\section{Experiments}

\subsection{Experimental Data}

The data in our study is drawn from the GlobalPhone  speech (GPS) database \cite{schultz2013globalphone}. GPS is a multilingual read speech resource where the utterances are recorded from native speakers (with self-reported gender labels) in a controlled recording environment with minimal noise. Therefore, the recording conditions across languages are comparable which enables us to conduct a cross-linguistic comparison. We experiment with seven Indo-European languages: Czech (\textsc{cze}), Polish (\textsc{pol}), Russian (\textsc{rus}), Bulgarian (\textsc{bul}), Brazilian Portuguese (\textsc{por}), French (\textsc{fra}), and German (\textsc{deu}). We acknowledge that our language sample is not typologically diverse. However, one of our objectives in this paper is to investigate whether the cross-lingual similarity of AWEs can predict the degree of mutual intelligibility between related languages. Therefore, we focus on the Slavic languages in this sample (\textsc{cze}, \textsc{pol}, \textsc{rus}, and \textsc{bul}), which are known to be typologically similar and mutually intelligible to various degrees. 

To train our AWE models, we obtain time-aligned spoken-word segments using the Montreal Forced Aligner \cite{mcauliffe2017montreal}. Then, we sample 42 speakers of balanced gender from each language. For each language, we sample \textasciitilde 32k spoken-word segments that are longer than 3 phonemes in length and shorter than 1.1 seconds in duration  (see Table~\ref{tab:data-stats} in Appendix \ref{sec:appendix_a} for word-level summary statistics of the data). For each word type, we obtain an IPA transcription using the grapheme-to-phoneme (G2P) module of the automatic speech synthesizer, \textit{eSpeak}. Each acoustic word segment is parametrized  as a sequence of 39-dimensional Mel-frequency spectral coefficients where frames are extracted over intervals of 25ms with 10ms overlap.

\subsection{Architecture and Hyperparameters}

\noindent
\textbf{Acoustic Encoder} \hspace{0.05cm} We employ a 2-layer recurrent neural network with a bidirectional Gated Recurrent Unit (BGRU) of hidden state dimension of 512, which yields  a 1024-dimensional AWE.  \vspace{0.1cm}

\noindent
\textbf{Training Details} \hspace{0.05cm} All models in this study are trained for 100 epochs with a batch size of 256  using the ADAM optimizer \cite{DBLP:journals/corr/KingmaB14} and an initial learning rate (LR) of 0.001. The LR is reduced by a factor of 0.5 if the mean average precision (mAP) for word discrimination on the validation set does not improve for 10 epochs. The epoch with the best validation performance during training is used for evaluation on the test set. \vspace{0.1cm}

\noindent
\textbf{Implementation} \hspace{0.05cm}  We build our models using PyTorch \cite{paszke2019pytorch}  and use FAISS \cite{JDH17} for efficient similarity search during evaluation. Our code is publicly available on GitHub.\footnote{\url{https://github.com/uds-lsv/xRSA-AWEs}}

\subsection{Quantitative Evaluation}
We evaluate the models using the standard intrinsic evaluation of AWEs: the same-different word discrimination task \cite{carlin2011rapid}. This task aims to assess the ability of a model to determine whether or not two given speech segments correspond to the same word type, which is quantified using a retrieval metric (mAP) reported in Table~\ref{tab:mAP}.

\begin{table}[t]
\centering
\begin{tabular}{c|ccc}
\hline
\multicolumn{1}{c|}{} & \multicolumn{3}{c}{Encoder type}     \\
\multicolumn{1}{c|}{\multirow{-2}{*}{Language }} & 
  \cellcolor[HTML]{e0e0e0}\small{PGE} &
  \cellcolor[HTML]{e0e0e0}\small{CAE} &
  \cellcolor[HTML]{e0e0e0}\small{CSE} \\ \hline 
\textsc{cze}         & 78.3 & 76.1 & \textbf{82.9} \\
\textsc{pol}         & 67.6 & 63.5 & \textbf{73.8} \\
\textsc{rus}         & 64.3 & 57.7 & \textbf{71.0} \\
\textsc{bul}         & 72.1 & 68.9 & \textbf{78.4} \\
\textsc{por}         & 74.5 & 72.2 & \textbf{80.4} \\
\textsc{fra}         & 65.6 & 64.5 & \textbf{68.5} \\
\textsc{deu}         & 67.9 & 70.3 & \textbf{75.8} \\ \hline
\end{tabular}
\caption{mAP performance on evaluation sets. }
\label{tab:mAP}
\end{table}

\section{Representational Similarity Analysis}
\begin{figure*}[ht]
    \centering
    \includegraphics[width=1.0\textwidth]{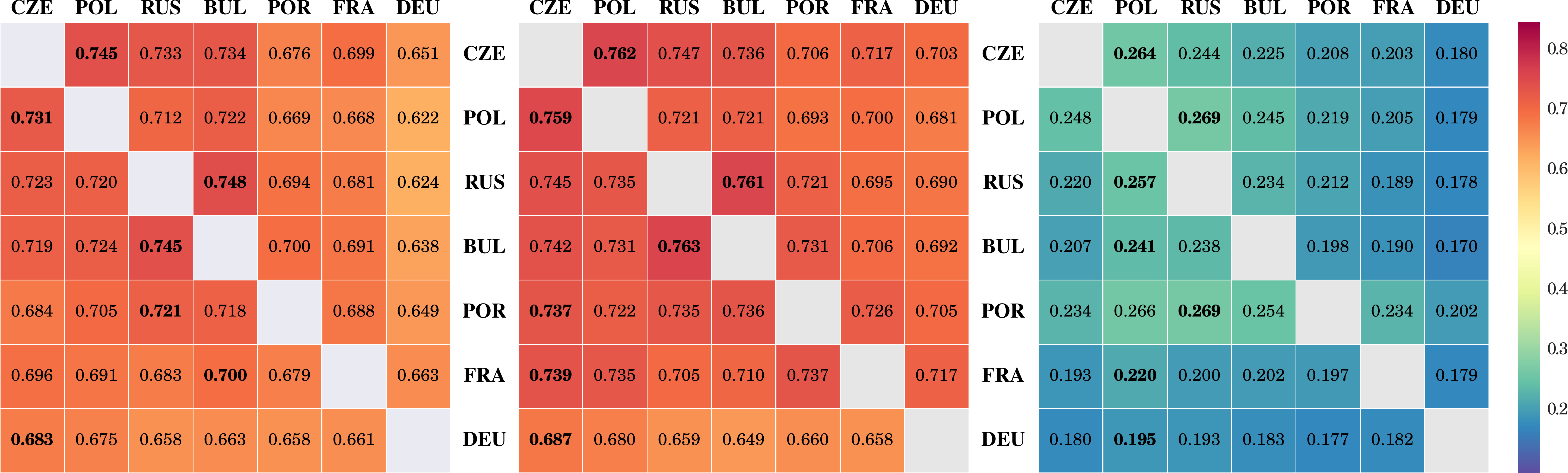}
    \caption{The cross-lingual representational similarity matrix (xRSM) for each model: PGE (Left), CAE (Middle), and CSE (Right). Each row corresponds to the language of the spoken-word stimuli while each column corresponds to the language of the encoder. Note that the matrices are not symmetric. For example in the PGE model, the cell at row \textsc{cze} and column \textsc{pol} holds the value of $\textsf{sim}(\textsc{cze}, \textsc{pol}) = \textsf{CKA}(\mathbf{X}^{(\textsc{cze}/\textsc{cze})}, \mathbf{X}^{(\textsc{cze}/\textsc{pol})}) = 0.745$, while the cell at row \textsc{pol} and column \textsc{cze} holds the value of $\textsf{sim}(\textsc{pol}, \textsc{cze}) = \textsf{CKA}(\mathbf{X}^{(\textsc{pol}/\textsc{pol})}, \mathbf{X}^{(\textsc{pol}/\textsc{cze})}) = 0.731$. }
    \label{fig:RSA_matrix}
\end{figure*}
Figure~\ref{fig:RSA_matrix} shows the cross-lingual representational similarity matrices (xRSMs) across the three different models using the linear CKA similarity metric.\footnote{The xRSMs obtained using non-linear CKA with an RBF kernel are shown in Figure~\ref{fig:xRSM_matrix_non_linear} in Appendix~\ref{sec:appendix_b}, which show very similar trends to those observed in Figure~\ref{fig:RSA_matrix}. } Warmer colors indicate a higher similarity between two representational spaces. One can observe striking differences between the PGE-CAE models on the one hand, and the CSE model on the other hand. The PGE-CAE models yield representations that are cross-lingually more similar to each other compared to those obtained from the contrastive CSE model. For example, the highest similarity score from the PGE model is $\textsf{sim}(\textsc{rus}, \textsc{bul}) = 0.748$, which means that the representations of the Russian word stimuli from the Bulgarian model $\mathcal{F}^{(\textsc{bul})}$ exhibit the highest representational similarity to the representations of the native Russian model $\mathcal{F}^{(\textsc{rus})}$. On the other hand, the lowest similarity score from the PGE model is $\textsf{sim}(\textsc{pol}, \textsc{deu}) = 0.622$, which shows that the German model's view of the Polish word stimuli is the view that differs the most from the view of the native Polish model. Likewise, the highest similarity score we observe from the CAE model is $\textsf{sim}(\textsc{bul}, \textsc{rus}) = 0.763$, while the lowest similarity is $\textsf{sim}(\textsc{deu}, \textsc{bul}) = 0.649$. If we compare these scores to those of the contrastive CSE model, we observe significant cross-language differences as the highest similarity scores are $\textsf{sim}(\textsc{pol}, \textsc{rus}) = \textsf{sim}(\textsc{por}, \textsc{rus}) = 0.269$, while the lowest is $\textsf{sim}(\textsc{bul}, \textsc{deu}) = 0.170$. This discrepancy between the contrastive model and the other models suggests that training AWE models with a contrastive objective hinders the ability of the encoder to learn high-level phonological abstractions and therefore contrastive encoders are more sensitive to the cross-lingual variation during inference compared to their sequence-to-sequence counterparts. 

\subsection{Cross-Lingual Comparison}
\begin{figure*}[t]
    \centering
    \includegraphics[width=0.80\textwidth]{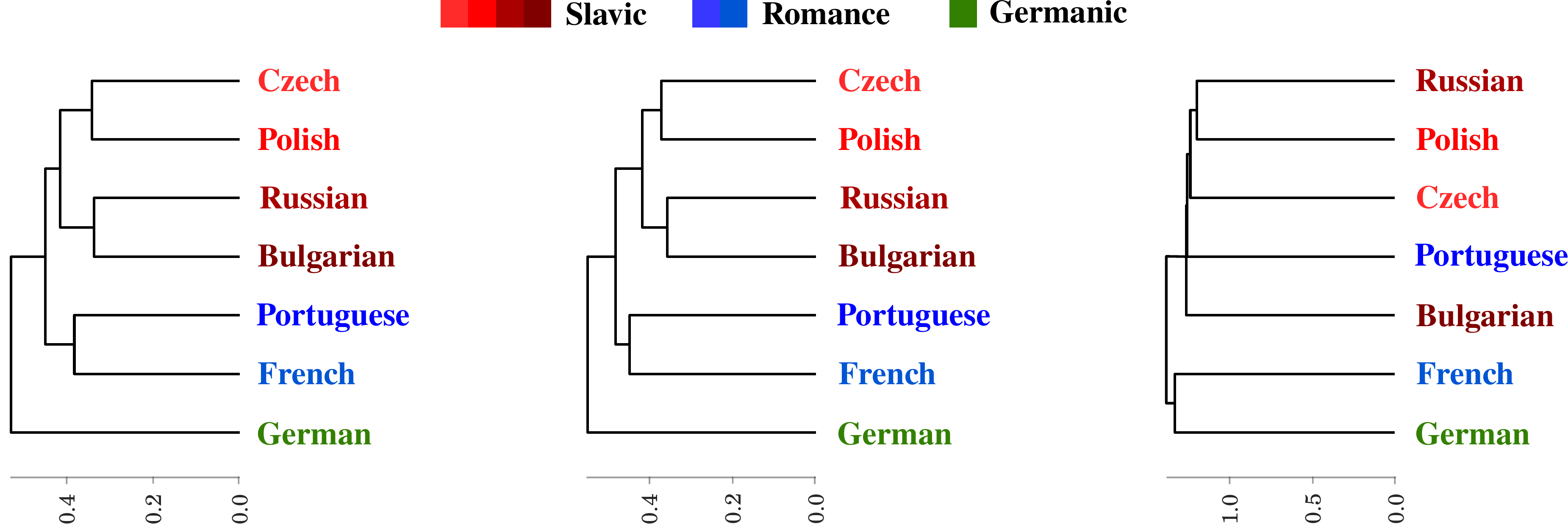}
    \caption{Hierarchical clustering analysis on the cross-lingual representational similarity matrices (using linear CKA) of the three models:  PGE (Left), CAE (Middle), and CSE (Right).}
    \label{fig:RSA_cluster}
\end{figure*}

To get further insights into how language similarity affects the model representations of non-native spoken-word segments, we apply hierarchical clustering on the xRSMs in Figure~\ref{fig:RSA_matrix} using the Ward algorithm \cite{ward1963hierarchical} with Euclidean distance. The result of the clustering analysis is shown in Figure~\ref{fig:RSA_cluster}. Surprisingly, the generated trees from the xRSMs of the PGE and CAE models are identical, which could indicate that these two models induce similar representations when trained on the same data. Diving a level deeper into the cluster structure of these two models, we observe that the Slavic languages form a pure cluster. This could be explained by the observation that some of the highest pair-wise similarity scores among the PGE models are observed between Russian and Bulgarian, i.e., $\textsf{sim}(\textsc{rus}, \textsc{bul}) = 0.748$ and $\textsf{sim}(\textsc{bul}, \textsc{rus}) = 0.745$, and Czech and Polish, i.e., $\textsf{sim}(\textsc{cze}, \textsc{pol}) = 0.745$. The same trend can be observed in the CAE models:  i.e., $\textsf{sim}(\textsc{rus}, \textsc{bul}) = 0.761$, $\textsf{sim}(\textsc{bul}, \textsc{rus}) = 0.763$, and $\textsf{sim}(\textsc{cze}, \textsc{pol}) = 0.762$. Within the Slavic cluster, the West Slavic languages Czech and Polish are grouped together, while the Russian (East Slavic) is first grouped with Bulgarian (South Slavic) before joining the West Slavic group to make the Slavic cluster. Although Russian and Bulgarian belong to two different Slavic branches, we observe that this pair forms the first sub-cluster in both trees, at a distance smaller than that of the West Slavic cluster (Czech and Polish). At first, this might seem surprising as we would expect the West Slavic cluster to be formed at a lower distance given the similarities among the West Slavic languages which facilitates cross-language speech comprehension, as documented by sociolinguistic studies \citep{golubovic2016mutual}. However, Russian and Bulgarian share typological features at the acoustic-phonetic and phonological levels which distinguish them from West Slavic languages.\footnote{Note that our models do not have access to word orthography. Thus, the similarity cannot be due to the fact that Russian and Bulgarian use Cyrillic script.} We further elaborate on these typological features in \S\ref{sec:discussion}. Even though Portuguese was grouped with French in the cluster analysis, which one might expect given that both are Romance languages descended from Latin, it is worth pointing out that the representations of the Portuguese word stimuli from the Slavic models show a higher similarity to the representations of the native Portuguese model compared to these obtained from the French model (with only two exceptions, the Czech PGE model and Polish CAE model).  We also provide an explanation of why this might be the case  in \S\ref{sec:discussion}.

The generated language cluster from the CSE  model does not show any clear internal structure with respect to language groups since all cluster pairs are grouped at a much higher distance compared to the PGE and CAE models. Furthermore, the Slavic languages in the generated tree do not form a pure cluster since Portuguese was placed inside the Slavic cluster. We also do not observe the West Slavic group as in the other two models since Polish was grouped first with Russian, and not Czech. We believe that this unexpected behavior of the CSE models can be related to the previously attributed to the poor performance of the contrastive AWEs in capturing word-form similarity \cite{abdullah21_interspeech}. 

Moreover, it is interesting to observe that German seems to be the most distant language to the other languages in our study. This observation holds across all three encoders since the representations of the German word stimuli by non-native models are the most dissimilar compared to the representations of the native German  model.

\subsection{Cross-Model Comparison}

\begin{figure}[t]
    \centering
    \includegraphics[width=0.50\textwidth]{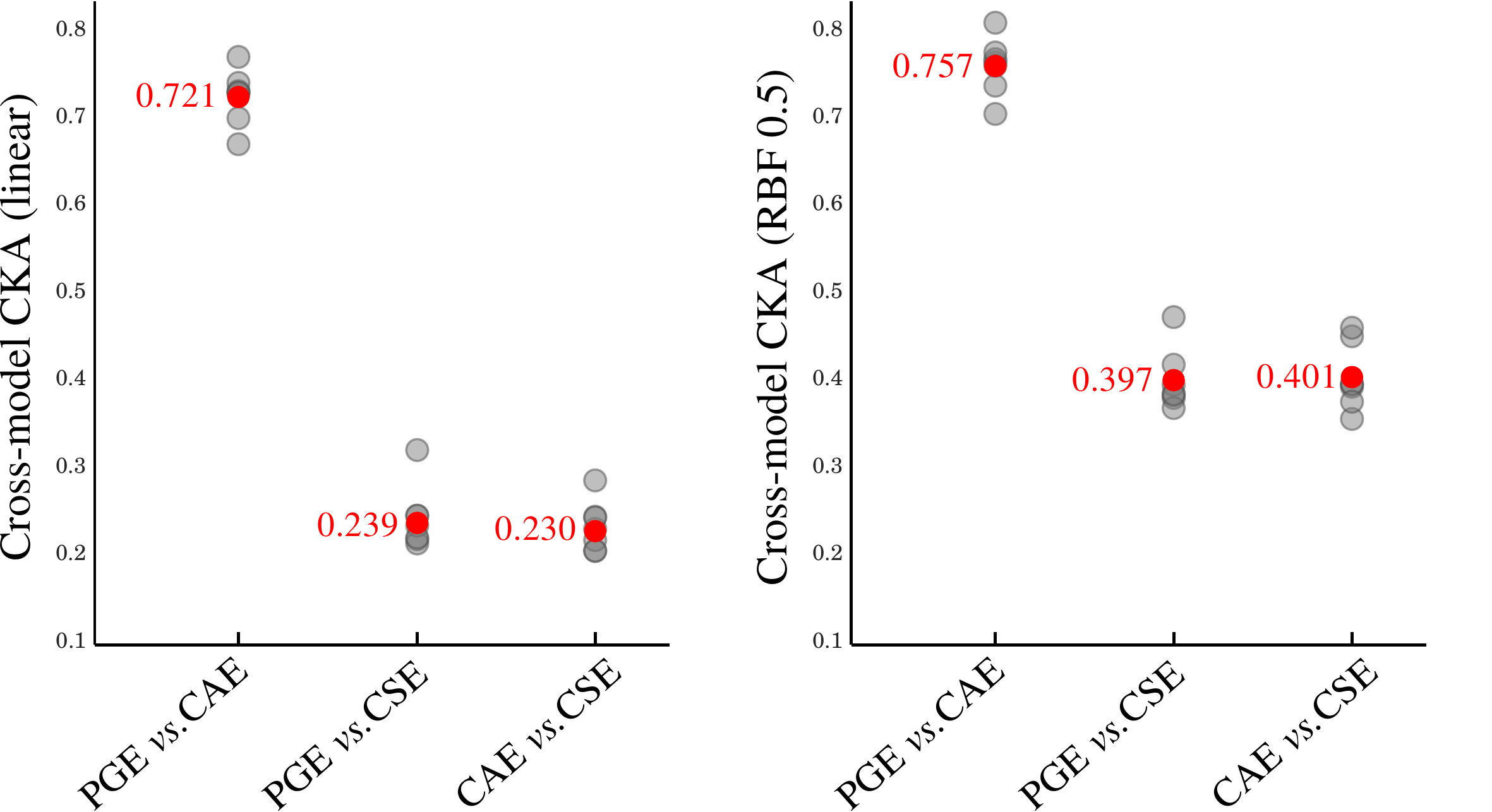}
    \caption{Cross-model CKA scores: linear CKA (left) and non-linear CKA (right). Each point in this plot is the within-language representational similarity of two models trained with different objectives when tested on the same (monolingual) word stimuli. Each red point is the average CKA score per model pair.}
    \label{fig:RSA_cross_model}
\end{figure}

To verify our hypothesis that the models trained with decoding objectives (PGE and CAE) learn similar representations when trained on the same data, we conduct a similarity analysis between the models in a setting where we obtain views of the spoken-word stimuli from native models, then compare these views across different learning objectives while keeping the language fixed using CKA as we do in our cross-lingual comparison. The results of this analysis is shown in Figure~\ref{fig:RSA_cross_model}. We observe that across different languages the pairwise representational similarity of the PGE-CAE models (linear $\textsf{CKA} = 0.721$) is very high in comparison to that of PGE-CSE models (linear $\textsf{CKA} = 0.239$) and CAE-CSE models (linear $\textsf{CKA} = 0.230$).\footnote{\textsf{CKA} scores are averaged over languages.} Although the non-linear CKA similarity scores tend to be higher, the general trend remains identical in both measures. This finding validates our hypothesis that the PGE and CAE models yield similar representation spaces when trained on the same data even though their decoding objectives operate over different modalities. That is, the decoder of the PGE model aims to generate the word's phonological structure in the form of a sequence of discrete phonological units, while the decoder of the CAE model aims to generate an instance of the same word represented as a sequence of (continuous) spectral vectors. Moreover, the sequences that these decoders aim to generate vary in length (the mean phonological sequence length is \textasciitilde 6 phonemes while mean spectral sequence length is \textasciitilde 50 vectors). Although the CAE model has no access to abstract phonological information of the word-forms it is trained on, it seems that this model learns non-trivial knowledge about word phonological structure as demonstrated by the representational similarity of its embeddings to those of a word embedding model that has access to word phonology (i.e., PGE) across different languages.

\section{Discussion}
\label{sec:discussion}
\subsection{Relevance to Related Work} 
Although the idea of representational similarity analysis (RSA) has originated in the neuroscience literature \citep{kriegeskorte2008representational}, researchers in NLP and speech technology have employed a similar set of techniques to analyze emergent representations of multi-layer neural networks. For example, RSA has previously been employed to analyze the correlation between neural and symbolic representations \citep{chrupala-alishahi-2019-correlating}, contextualized word representations \citep{abnar-etal-2019-blackbox, abdou-etal-2019-higher, lepori-mccoy-2020-picking, wu2020similarity}, representations of self-supervised speech models \cite{chung2021similarity} and visually-grounded speech models \citep{chrupala-etal-2020-analyzing}. We take the inspiration from previous work and apply RSA in a unique setting: to analyze the impact of typological similarity between languages on the representational geometry. Our analysis focuses on neural models of spoken-word processing that are trained on naturalistic speech corpora. Our goal is two-fold: (1) to investigate whether or not neural networks exhibit predictable behavior when tested on speech from a different language (L2), and (2) to examine the extent to which the training strategy affects the emergent representations of L2 spoken-word stimuli. To the best of our knowledge, our study is the first to analyze the similarity of emergent representations in neural acoustic models from a cross-linguistic perspective.

\subsection{A Typological Discussion}

Given the cross-linguistic nature of our study, we choose to discuss our findings on the representational similarity analysis from a language typology point of view. Language typology is a sub-field within linguistics that is concerned with the  study  and  categorization  of  the  world’s  languages based on their linguistic structural properties \cite{comrie1988linguistic, crofttypology}. Since acoustic word embeddings are word-level representations induced from actual acoustic realizations of word-forms, we focus on the phonetic and phonological properties of the languages in our study. We emphasize that our goal is not to use the method we propose in this paper as a computational approach for phylogenetic reconstruction, that is, discovering the historical relations between languages. However, typological similarity usually correlates with phylogenetic distance since languages that have diverged from a common historical ancestor inherited most of their typological features from the ancestor language (see \citet{bjerva2019language} for a discussion on how typological similarity and genetic relationships interact when analyzing neural embeddings). 

Within the language sample we study in this paper, four of these languages belong to the Slavic branch of Indo-European languages. Compared to the Romance and Germanic branches of Indo-European, Slavic languages are known to be remarkably more similar to each other not only at lexical and syntactic levels, but also at the pre-lexical level (acoustic-phonetic and phonological features). These cross-linguistically shared features between Slavic languages include rich consonant inventories, phonemic iotation and complex consonant clusters. The similarities at different linguistic levels facilitate spoken intercomprehension, i.e., the ability of a listener to comprehend an utterance (to some degree) in a language that is unknown, but related to their mother tongue. Several sociolinguistic studies of mutual intelligibility have reported a higher degree of intercomprehension among speakers of Slavic languages compared to other language groups in Europe \cite{golubovic2016mutual, gooskens2018mutual}. On the other hand, and even though French and Portuguese are both Romance languages, they are less mutually intelligible compared to Slavic language pairs as they have diverged in their lexicons and phonological structures \cite{gooskens2018mutual}. This shows that cross-language speech intelligibility is not mainly driven by shared historical relationships, but by contemporary typological similarities. 

Therefore, and given the documented phonological similarities between Slavic languages, it is not surprising that Slavic languages form a pure cluster in our clustering analysis over the xRSMs of two of the models we investigate. Furthermore, the grouping of Czech-Polish and Russian-Bulgarian in the Slavic cluster can be explained if we consider word-specific suprasegmental features. Besides the fact that Czech and Polish are both West Slavic languages that form a spatial continuum of language variation, both languages have fixed stress. The word stress in Czech is always on the initial syllable, while Polish has penultimate stress, that is, stress falls on the syllable preceding the last syllable of the word \cite{sussex2006slavic}. On the other hand, Russian and Bulgarian languages belong to different Slavic sub-groups---Russian is East Slavic while Bulgarian is South Slavic. The representational similarity between Russian and Bulgarian can be attributed to the typological similarities between Bulgarian and East Slavic languages. In contrast to West Slavic languages, the stress in  Russian and Bulgarian is free (it can occur on any syllable in the word) and movable (it moves between syllables within morphological paradigms). Slavic languages with free stress tend to have a stronger contrast between stressed and unstressed syllables, and vowels in unstressed syllables undergo a process that is known as vowel-quality alternation, or lexicalized vowel reduction \citep{barry2001cross}. For example, consider the Russian word \textit{ruka} (`hand'). In the singular nominative case the word-form is phonetically realized as \textit{ruk\'a} \textipa{[r\textupsilon\textprimstress ka]} while in the singular accusative case \textit{r\'uku} \textipa{[\textprimstress ruk\textupsilon]}  \cite{sussex2006slavic}. Here, the unstressed high back vowel \textipa{/u/} is realized as \textipa{[\textupsilon]} in both word-forms. Although vowel-quality alternations in Bulgarian follow different patterns, Russian and Bulgarian vowels in unstressed syllables are reduced in temporal duration  and quality. Therefore, the high representational similarity between Russian and Bulgarian could be explained if we consider their typological similarities in word-specific phonetics and prosody.

The Portuguese language presents us with an interesting case study of language variation. From a phylogenetic point of view, Portuguese and French have diverged from Latin and therefore they are both categorized as Romance languages. The clustering of the xRSMs of the PGE and CAE models groups Portuguese and French together at a higher distance compared to the Slavic group, while Portuguese was grouped with Slavic languages when analyzing the representations of the contrastive CSE model. Similar to Russian and Polish, sibilant consonants (e.g., \textipa{/S/, /Z/}) and the velarized (dark)  \textit{L}-sound (i.e., /\textipa{\textltilde}/) are frequent in Portuguese. We hypothesize that contrastive training encourages the model to pay more attention to the segmental information (i.e., individual phones) in the speech signal at the expense of phonotactics (i.e., phone sequences). Given that contrastive learning is a predominant paradigm in speech representation learning \citep{Oord2018RepresentationLW, schneider19_interspeech}, we encourage further research to analyze whether or not speech processing models trained with contrastive objectives exhibit a similar behavior to that observed in human listeners and closely examine their plausibility for cognitive modeling.

\subsection{Implications on Cross-Lingual Transfer Learning}

A recent study on zero-resource AWEs has shown that cross-lingual transfer is more successful when the source (L1) and target (L2) languages are more related \cite{jacobs21_interspeech}. We conducted a preliminary experiment on the cross-lingual word discrimination performance of the models in our study and observed a similar effect. However, we also observed that cross-lingual transfer using the contrastive model is less effective compared to the models trained with decoding objectives. From our reported RSA analysis, we showed that the contrastive objective yields models that are cross-lingually dissimilar compared to the other objectives. Therefore, future work could investigate the degree to which our proposed RSA approach predicts the effectiveness of different models in a cross-lingual zero-shot scenario.

\section{Conclusion}
We presented an experimental design based on representational similarity analysis (RSA) whereby we analyzed the impact of language similarity on representational similarity of acoustic word embeddings (AWEs). Our experiments have shown that AWE models trained using decoding objectives exhibit a higher degree of representational similarity if their training languages are typologically similar. We discussed our findings from a typological perspective and highlighted pre-lexical features that could have an impact on the models' representational geometry. Our findings provide evidence that AWE models can predict the facilitatory effect of language similarity on cross-language speech perception and complement ongoing efforts in the community to assess their utility in cognitive modeling. Our work can be further extended by considering speech segments below the word-level (e.g., syllables, phonemes), incorporating semantic representations into the learning procedure, and investigating other neural architectures. 

\section*{Acknowledgements}
We thank the anonymous reviewers for their constructive comments and insightful feedback. We further extend our gratitude to Miriam Schulz and Marius Mosbach for proofreading the paper. This research is funded by the Deutsche Forschungsgemeinschaft (DFG, German Research Foundation), Project ID 232722074 -- SFB 1102.

\bibliography{anthology,custom}
\bibliographystyle{acl_natbib}

\newpage
\appendix

\section*{Appendices}

\section{Experimental Data Statistics} 
\label{sec:appendix_a}

Table \ref{tab:data-stats} shows a word-level summary statistics of our experimental data extracted from the GlobalPhone speech daabase (GPS).

\begin{table*}[h]
\centering
\begin{tabular}{cccccccc}
\hline
 \begin{tabular}[c]{@{}c@{}}Lang. \end{tabular} &
  \begin{tabular}[c]{@{}c@{}}Language\\ group\end{tabular} &
  \begin{tabular}[c]{@{}c@{}}\#Train\\ spkrs\end{tabular} &
  \begin{tabular}[c]{@{}c@{}}\#Train \\ samples\end{tabular} &
  \begin{tabular}[c]{@{}c@{}}\#Eval \\ samples\end{tabular} &
  \begin{tabular}[c]{@{}c@{}}\#Phones/word\\ (mean $\pm$ SD)\end{tabular} &
  \begin{tabular}[c]{@{}c@{}} Word dur. (sec)\\ (mean $\pm$ SD)\end{tabular} &
  \begin{tabular}[c]{@{}c@{}}Token-Type\\ ratio\end{tabular} \\ \hline
\rowcolor[HTML]{e0e0e0}\textsc{cze}  & West Slavic     & 42 & 32,063 & 9,228 & 6.77 $\pm$ 2.28 & 0.492 $\pm$ 0.176   & 0.175 \\
\textsc{pol}  & West Slavic     & 42 & 31,507 & 9,709 & 6.77 $\pm$ 2.27 & 0.488 $\pm$ 0.175 & 0.192 \\
\rowcolor[HTML]{e0e0e0}\textsc{rus}  & East Slavic     & 42 & 31,892 & 9,005 & 7.56 $\pm$ 2.77 & 0.496 $\pm$ 0.163 & 0.223 \\
\textsc{bul}  & South Slavic    & 42 & 31,866 & 9,063 & 7.24 $\pm$ 2.60 & 0.510 $\pm$ 0.171   & 0.179 \\
\rowcolor[HTML]{e0e0e0}\textsc{por}  & Romance   & 42 & 32,164 & 9,393 & 6.95 $\pm$ 2.36 & 0.526 $\pm$ 0.190 & 0.134 \\
\textsc{fra}  & Romance & 42 & 32,497 & 9,656 & 6.24 $\pm$ 1.95 & 0.496 $\pm$ 0.163 & 0.167 \\
\rowcolor[HTML]{e0e0e0}\textsc{deu}  & Germanic   & 42 & 32,162 & 9,865 & 6.57 $\pm$ 2.47 & 0.435 $\pm$ 0.178 & 0.150 \\ \hline
\end{tabular}
\caption{Summary statistics of our experimental data.}
\label{tab:data-stats}
\end{table*}

\section{Representational Similarity with Non-Linear CKA} 
\label{sec:appendix_b}

We  provide the cross-lingual representational similarity matrices (xRSMs) constructed by applying the non-linear CKA measure with a radial basis function (RBF) in Figure \ref{fig:xRSM_matrix_non_linear} below. We observe similar trends to those presented in Figure~\ref{fig:RSA_matrix}. 

Applying hierarchical clustering on the xRSMs in Figure \ref{fig:xRSM_matrix_non_linear} yields the language clusters shown in Figure \ref{fig:RSA_cluster_non_linear}. We observe that the clusters are identical to those shown in Figure~\ref{fig:RSA_cluster}, except for the CSE model where Portuguese is no longer in the Slavic group. However, Portuguese remains the non-Slavic language that is the most similar to Slavic languages.

\begin{figure*}[h]
    \centering
    \includegraphics[width=1.0\textwidth]{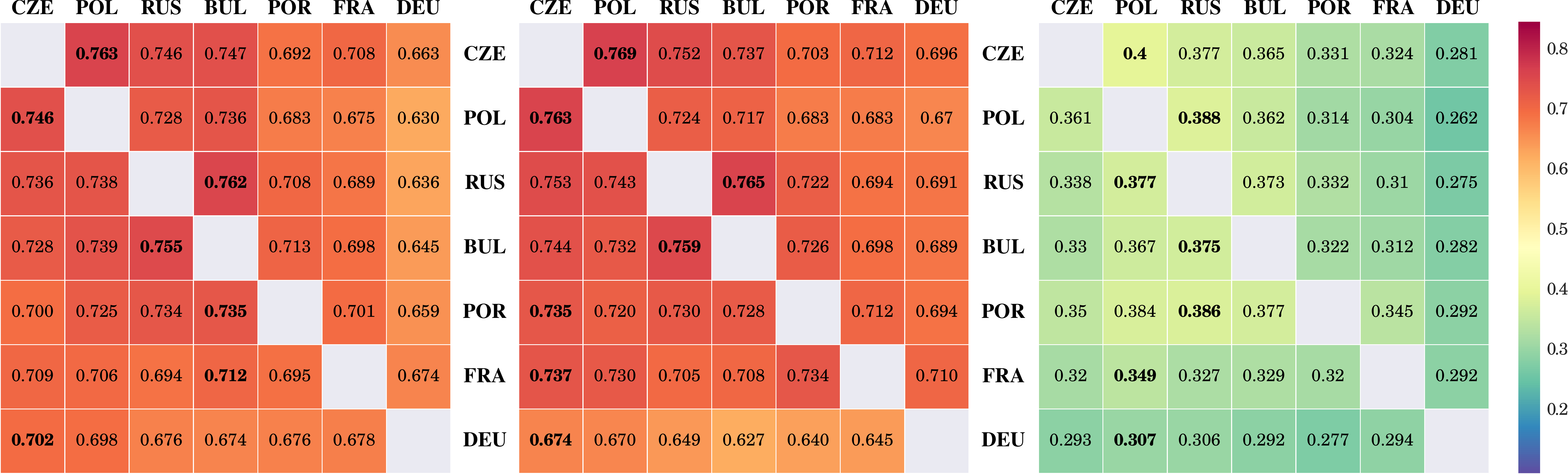}
    \caption{The cross-lingual representational similarity matrix (xRSM) for each model: PGE (Left), CAE (Middle), and CSE (Right), obtained using the non-linear CKA measure. Each row corresponds to the language of the spoken-word stimuli while each column corresponds to the language of the encoder. }
    \label{fig:xRSM_matrix_non_linear}
\end{figure*}

\begin{figure*}[h]
    \centering
    \includegraphics[width=0.80\textwidth]{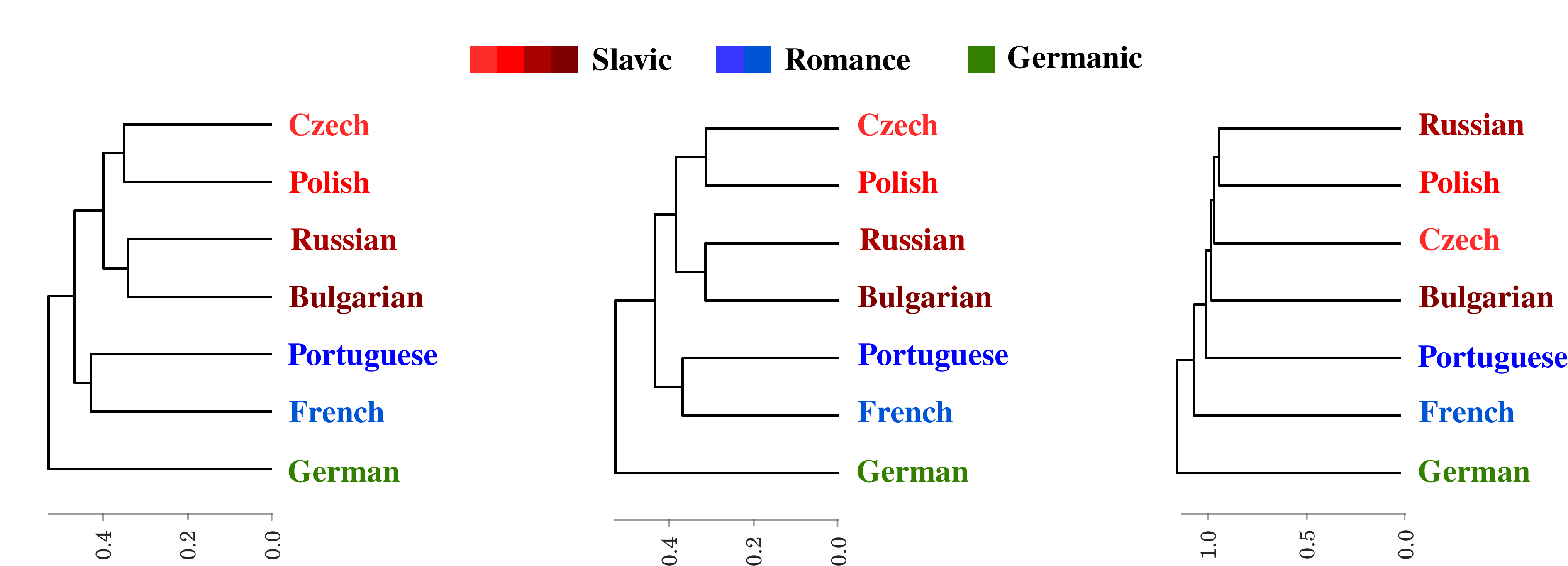}
    \caption{Hierarchical clustering analysis on the cross-lingual representational similarity matrices (using non-linear CKA) of the three models:  PGE (Left), CAE (Middle), and CSE (Right).}
    \label{fig:RSA_cluster_non_linear}
\end{figure*}

\end{document}